\documentclass[letterpaper, 9pt, conference]{ieeeconf}  

\IEEEoverridecommandlockouts

\usepackage{graphics} 
\usepackage{epsfig} 
\usepackage{mathptmx} 
\usepackage{times} 
\usepackage{amsmath} 
\usepackage{amssymb}  
\usepackage{graphicx}
\usepackage{bm}
\usepackage{eucal}

\usepackage{float}
\usepackage[ruled,vlined]{algorithm2e}
\usepackage{pgfplotstable}
\usepackage{pgfplots}
\usepackage{tikz}
\usetikzlibrary[pgfplots.groupplots]
\usepgfplotslibrary{fillbetween}
\pgfplotsset{compat=1.5}
\usepackage{subcaption}
\usepackage{caption}
\usepackage{xcolor}
\usepackage{enumerate}
\usepackage{makecell}
\usepackage{colortbl}

\usepackage{graphicx}
\usepackage{tabularx,ragged2e,booktabs}
\usepackage{url}
\usepackage{siunitx}
\usepackage{rotating}
\usepackage{amssymb}
\usepackage{tabularx}
\usepackage{pgfplots}
\usepackage{csvsimple}
\usepackage{multirow}
\usepackage{cite}
\usepackage{float}
\usepackage{graphicx}
\usepackage{booktabs}
\usepackage{todonotes}
\usepackage{multirow}
\usepackage{amsmath}
\usepackage{wrapfig}
\usepackage[font=footnotesize]{caption}
\SetKwComment{Comment}{$\triangleright$\ }{}

\SetCommentSty{mycommfont}
\DeclareMathOperator*{\argmax}{arg\,max}

\title{\LARGE \bf
Push to know! - Visuo-Tactile based Active Object Parameter Inference with Dual Differentiable Filtering   
}
\author{Anirvan Dutta, Etienne Burdet and Mohsen Kaboli
\thanks{A. Dutta and M. Kaboli are with the BMW Group, Munich, Germany. 
e-mail: name.surname@bmwgroup.com}%
\thanks{A. Dutta and E. Burdet are with Imperial College of Science, Technology and Medicine, London, UK. M. Kaboli is with Radboud University, Netherlands.}
\thanks{Funded in part by the EU H2020 INTUITIVE under Grant ID 861166 and in part by EU Horizon PHASTRAC under Grant ID 101092096.}
}

\begin{document}
\bstctlcite{IEEEexample:BSTcontrol}

\maketitle
\thispagestyle{empty}
\pagestyle{empty}

\begin{abstract}
For robotic systems to interact with objects in dynamic environments, it is essential to perceive the physical properties of the objects such as shape, friction coefficient, mass, center of mass, and inertia. This not only eases selecting manipulation action but also ensures the task is performed as desired. However, estimating the physical properties of especially novel objects is a challenging problem, using either vision or tactile sensing. In this work, we propose a novel framework to estimate key object parameters using non-prehensile manipulation using vision and tactile sensing.
Our proposed active dual differentiable filtering (ADDF) approach as part of our framework learns the object-robot interaction during non-prehensile object push to infer the object's parameters. Our proposed method enables the robotic system to employ vision and tactile information to interactively explore a novel object via non-prehensile object push. The novel proposed $N$-step active formulation within the differentiable filtering facilitates efficient learning of the object-robot interaction model and during inference by selecting the next best exploratory push actions (where to push? and how to push?). We extensively evaluated our framework in simulation and real-robotic scenarios, yielding superior performance to the state-of-the-art baseline. 
\end{abstract}

\section{Introduction}
\label{sec:introduction}
\begin{figure}[th!]
    \centering
    \includegraphics[width=0.9\columnwidth]{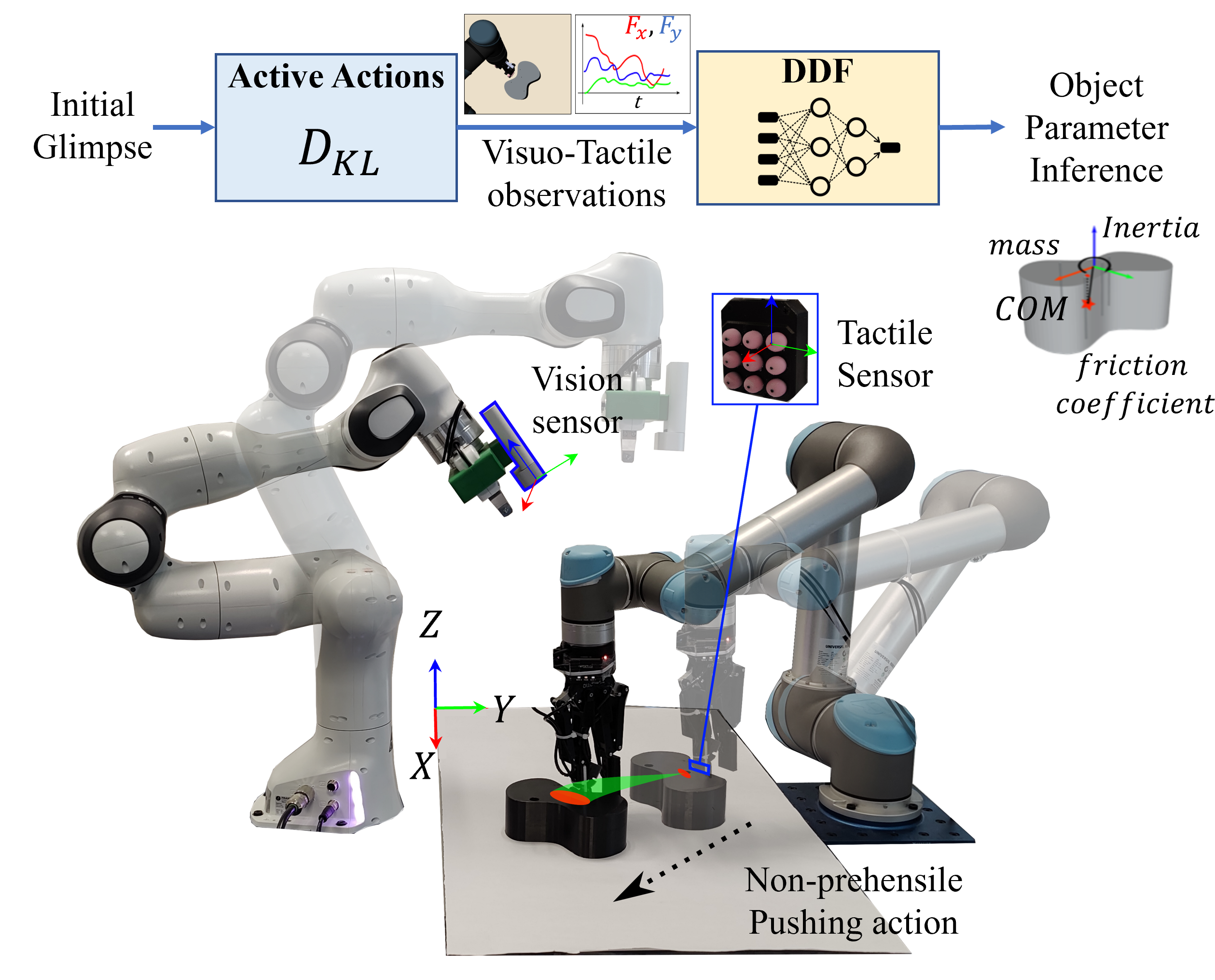}
    \caption{Problem setup for visuo-tactile based active object parameter inference}
    \label{fig:problem_setup}
\end{figure}

To manipulate novel objects, humans often perceive object properties through actions such as holding, grasping or pushing to gain better control \cite{etienne_3, kaboli2016tactile}. In such \textit{interactive visuo-tactile perception}, active physical interactions or explorations are made to enhance object perception ~\cite{bohg_17_interactive, bajcsy_active, seminara_active, kaboli_review_tro}. In this work, we investigate inferring physical object properties (such as friction coefficient, mass, center of mass, and inertia) with a robotic system using non-prehensile push actions via visuo-tactile sensory information (see Fig. \ref{fig:problem_setup}). 

Pushing an object to explore its parameters is simpler than grasping or lifting \cite{mason1999progress}, especially when the objects are large and heavy or when no prior knowledge about the object is present. Further, in grasping or lifting, object geometry, grasp stability and other factors come into play which is not the case for pushing. Moreover, since the object is not rigidly attached to the robotic end-effector, it exhibits a broader class of motions which can be used for parameter estimation \cite{mason1999progress}.

However, using non-prehensile pushing to infer object parameters is a challenging task as the interaction dynamics between the object and the robotic system are sophisticated to model, due to uncertainty in the contacts, surface irregularities etc. ~\cite{mason_pushing, push_survey}. This can make it difficult to infer the parameters accurately. Therefore the data-driven approach is a more viable approach for such an interaction model. 

To improve data efficiency and inference time, it is essential, the robotic systems, strategically select the next best exploratory push actions (active push exploration) such as where to push? and how to push? Authors have already shown that active object exploration outperforms uniform and random strategy for reducing the uncertainty about objects for problems like object recognition ~\cite{kaboli2019auro, kaboli2018active, kaboli2017tactile, feng2018active} and pose estimation ~\cite{murali_iros_21, murali_interactive}. In this work, we propose $N$-step Information Gain formulation for active exploratory push action selection which is crucial for dynamic interactions such as pushing. 

Furthermore, since physical properties of objects are not directly observable and must be inferred from noisy visuo-tactile sensory observations, we introduce a novel dual differentiable filtering approach to address this and effectively handle the time-invariant nature of such properties.

\section{Related Works}
\label{sec:relatedwork}
Estimating the physical properties of novel objects is a challenging problem in robotics, using either vision or tactile sensing. The physical object properties are not salient under static or quasi-static interactions, and often each parameter is only revealed under specific interactions, making it an interesting research problem \cite{dense_phys}.

One of the earliest works of Atkeson et al.~\cite{atkeson_86} estimated the mass and moment of inertia of an object rigidly attached to a manipulator, using joint torques and a wrist-mounted force-torque sensor. Similar results have also been presented in \cite{zhao_rel_inertia}. These approaches required the object to be manually attached to the end-effector. Few works elevated this constraint as in ~\cite{yu_pushe_05}, where authors used a 2-fingered bespoke mechanism to measure contact forces during planar pushing and in \cite{fukuda_pushe_99}, the authors applied a tilting approach to measure wrenches to estimate the inertial parameters. Zhao et al. \cite{zhao_param_18} incorporated friction estimation, by grasping the object and measuring the contact forces during the sliding regime. Most of these prior estimation techniques relied on precise force or tactile sensing, assumptions about the object geometry or the interaction between the object and environment, and employed specialized mechanisms, thus making it difficult for generalization and autonomous exploration of the object. 

Some researchers attempted to overcome the previously mentioned limitations by introducing interactive manipulation techniques like grasping or pushing.  In \cite{tanaka_rel_04}, the authors estimated only the mass of an object by controlled pushing, which required prior knowledge about the friction-co-efficient of the surface. Similarly, to determine the centre of mass of the object, Yao et al. \cite{kaboli_com_17} utilised the tactile forces during a 3-fingered robotic grasp. To estimate more object physical properties, Sundaralingam et al. \cite{bala_21} used a factor graph approach using in-hand manipulation with precise tactile and force-torque sensing. The approach relied on the approximation of in-hand object dynamics, known object shape and a marker-less tracking system. More recently, Uttayas et al. \cite{pakorn_etienne} estimated visco-elastic properties using filtering approach, relying on approximate spring mass damper model. The above-mentioned works employing interactive manipulation often used an analytical formulation to model the object-robot interaction, which is often approximate and has significant assumptions about the interactions. 

Recently, data-driven and physics engine approaches are being taken to overcome such problems. Wu et al. \cite{tenenbaum_data_15} used deep learning to learn interactions between objects colliding in a physics engine and utilised the learned model to estimate mass and friction parameters for real object motion. Song et al. \cite{song_19}, \cite{song_20} relied on a physics engine to predict expected object motions during pushing and employed Bayesian optimization on a real object motion to predict distributed mass and friction on objects. These works often relied on the accuracy of the physics engine and are generally computationally complex. Xu et al. \cite{dense_phys} used only vision and deep learning to learn a representation of the mass and friction coefficient by randomly pushing and poking objects. Mavrakis et al. \cite{nikos_base_work} collected large pushing trajectories (40k) in the simulation environment and learned a regression model for estimating an object's inertial parameters during non-prehensile pushing. However, these approaches require intensive training and do not involve strategic interaction. In this work, we propose an active formulation for efficient training data-driven object-robot interaction model.

As of yet, either vision or tactile were used to estimate the physical object properties. On one hand, tactile information is crucial to infer multiple object properties like in \cite{bala_21, kaboli_com_17, kaboli2015humanoids, kaboli2015hand}, however, requires precise position information and prior knowledge. On the other hand, vision-based approaches such as \cite{dense_phys, alina_1} could only estimate fewer object properties with higher error rates, but required no prior knowledge about the object. To exploit the complementing vision and tactile sensing modalities, we propose to utilise both. Recently, Murali et al. ~\cite{murali_interactive} and Lee et al. ~\cite{vt_lee_2021} have shown visuo-tactile based approach significantly improves the performance of the robotics systems problems like pose estimation and contact-rich manipulation. 

To tackle the above-aforementioned problems and constraints for estimating or inferring the physical object properties, we propose visuo-tactile based active object parameter inference with a dual differentiable filter. Our proposed approach enables the robotic system to utilise vision and tactile information to interactively explore a novel object via non-prehensile pushing, not requiring any prior knowledge about the object's shape, position etc. In addition, the novel $N$-step active formulation within the differentiable filtering facilitate efficient learning of the object-robot interaction model and also during inference by selecting the next best exploratory push actions. 

\subsection*{Contributions}
The contributions of the work are summarized in three folds.

\begin{enumerate}
    \item  We propose a novel dual differentiable filter to systematically account for the time-invariant nature of object parameters and time-varying object pose during non-prehensile pushing. 
    \item We propose a novel data-driven model for the dual differentiable filter which handles raw visuo-tactile sensory information and captures the object-robot interaction during pushing.
    \item We propose a novel $N-$step lookahead formulation exploiting the differentiable filtering step to select the next best exploratory push actions for both learning the object-robot interaction model and performing inference.
\end{enumerate}
We perform extensive simulation and real-robot experiments supporting the proposed method and comparison with the state-of-the-art baselines.

\section{Proposed Method}
\begin{figure*}[t!]
    \centering
    \includegraphics[width = 0.9\textwidth, height = 105mm]{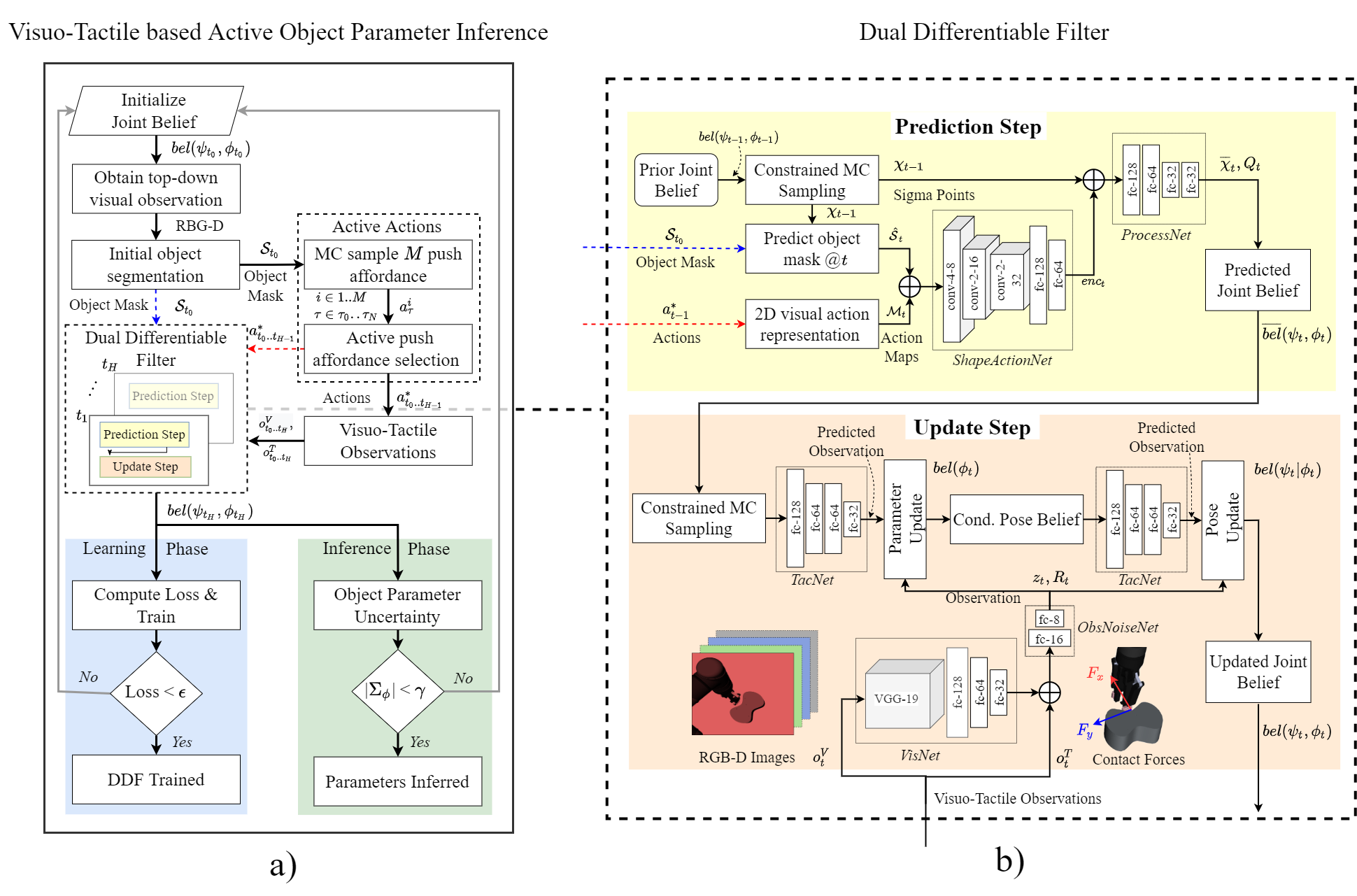}
    \caption{Our proposed framework (ADDF) for visuo-tactile based active object exploration using non-prehensile manipulation. Part a) presents the overall framework and part b) presents an expanded view of the dual differentiable filter block. }
    \label{fig:framework}
\end{figure*}

\subsection{Problem Definition}

In this work, we consider the problem of estimating the state $s$  of an unknown rigid object from vision $o^V$ and tactile observation $o^T$ using non-prehensile pushing actions $a$. At any given time $t$, the state $s_t = \{ \psi_t, \phi \}$ comprises of time-varying factors: \textbf{pose}, $\psi_t=\{x_t, y_t, \theta_t\}$ as well as time-invariant factors: \textbf{parameters},  $\phi= \{m, \mu, CoM_x, CoM_y, I_z \}$ as \textit{mass, relative friction coefficient between object and the surface, center of mass, inertia}. The center of mass is measured w.r.t frame attached to the geometric center of an object and only the rotational inertia in 2D is considered, as the interaction is restricted to motion in 2D.  The observation $o^V_t$ consists of RGB-D images of the pushing area and tactile observation $o^T_t$ comprising of \textit{2D contact forces, contact indicator}. The contact indicator $\in {0, 1}$, depending on whether the robot and object are in contact. The pushing action $a_t$ is parameterized by \textit{contact point $(cp)$}, \textit{push direction} $(pd)$ and \textit{velocity} $(v)$ of the push. The $cp$ comprises of 2D world co-ordinate of the contact point, $pd$, z-axis rotational angle of the robotic system aligned along a pushing direction \& $v$ is the magnitude of push velocity by the robotic system.

We perform quasi-static pushing \cite{mason_pushing} to infer the object parameters $\phi$ which are not directly observable either through vision or tactile sensing. Our proposed framework is illustrated in Fig.\ref{fig:framework} a). It comprises a novel dual differentiable filter for parameter and pose estimation along with data-driven models. The action selection for push affordance is done via computing $N$-step information gain term, making it an active dual differentiable filter (ADDF). Firstly, the robotic system learns data-driven models utilised within the differentiable filter. After learning, we perform inference on novel and unknown objects to estimate their parameters, with no prior information about the novel object. In the following sections, we explain the various components of the framework.

\subsection{Bayesian inference and Differential Filters}
We represent the belief about the current state of the object $s_t$ with a distribution conditioned on previous actions $a_{1:t}$ and observations $o_{1:t}$. This distribution is denoted as the belief of the state $bel(s_t)$  
\begin{equation}
     bel(s_t) = p(s_t|o_{1:t}, a_{1:t}) = \frac{p(o_t|s_t,o_{1:t-1}, a_{1:t})p(s_t|o_{1:t-1}, a_{1:t})}{p(o_t|o_{1:t-1},a_{1:t})}
     \label{eq:belief_state}
 \end{equation}
One prominent approach to computing the belief tractably is to employ Recursive Bayesian Filters which follow the structure as:
\begin{align}
    bel(s_t) = p(s_t|o_{1:t}, a_{1:t}) = \eta p(o_t|s_t,a_t)\overline{bel}(s_t)\\
    \overline{bel}(s_t) = \int_{}^{}p(s_t|s_{t-1}, a_{t-1})bel(s_{t-1})ds_{t-1} 
    \label{eq:bayes_filter}
\end{align}
Kalman Filters are a common choice of Bayesian Filtering which is optimal in linear systems and can be extended in non-linear cases through various approaches. Two key aspects of Bayesian filtering are the representation of the process model of the state in the form of $p(s_t|s_{t-1}, a_{t-1})$ and an observation likelihood model relating the states to the observations $p(o_t|s_t)$. For our problem, we employ a data-driven approach to learn the process and observation model along with the respective noise models, end-to-end using a differentiable filter. Recently, differentiable filters integrating Bayesian filtering with deep learning \cite{oli_df_3, alberto_push_1, lambert_push_2, levine_df_3} were proposed. The authors have also shown that such an approach performs better compared to the standard deep-learning approach in handling real-world noise and in \cite{alina_2_main} showed the strength of such approach for a variety of tasks like visual optometry, visual object tracking etc. 

In our problem, the data-driven models within a differentiable filter capture the complex and stochastic object-robot interaction model during non-prehensile pushing. Further, the pose of the object is intricately dependent on the parameters and straightforward combined (joint) filtering for pose and parameter does not perform well. We empirically show this in Section~\ref{sec:results}. Therefore, we utilize a dual filter design, exploiting the dependency among the states for consistent filtering and inferring the parameters of the object.  

\subsection{Dual Differentiable Filter}
We derive our dual filter based on differentiable UKF \cite{prob_rob_book, alina_2_main}. For the dual filter formulation, we explicitly represent the state $s_t$ by the joint distribution of $\psi_t$ and $\phi_t$, via Multivariate Gaussian distribution: 
\begin{equation}
     bel(\psi_t, \phi_t) \sim \mathcal{N}(\psi_t, \phi_t | \mu_t , \Sigma_t)
     \label{eq:joint_distribution}
 \end{equation}
with statistics $\mu_t \in \mathbb{R}^8$ and $\Sigma_t \in \mathbb{R}^{8x8}$ as 
\begin{align}
    \mu_t = \begin{pmatrix}
\mu_{\psi_t} \\
\mu_{\phi_t}
\end{pmatrix}, \quad \Sigma_t = \begin{pmatrix}
\Sigma_{\psi_t} & \Sigma_{{\psi_t\phi_t}}\\
\Sigma_{{\phi_t\psi}_t} & \Sigma_{\phi_t}
\end{pmatrix} \quad .
\end{align}
The dual filter as shown in Fig.\ref{fig:framework}(b) follows, the structure of a Kalman Filter with a \textit{prediction step} and an \textit{update step}, with key novelty explained in this section. 

\subsubsection{Prediction Step}
In prediction step, the next step joint belief is predicted given the prior belief and the actions. The object parameters are real physical quantities with some physical constraints, (for e.g. $m, \mu > 0$). However, simply constraining the sigma points $\chi^{UT}$ in UKF approach does not preserve the true variance of the Gaussian distribution \cite{ut_constraints}. Therefore, we perform constrained Monte-Carlo sigma point sampling to preserve the physical constraints and the variance of the Gaussian. We employ a differentiable sampling method \cite{mc_method} to sample $C$ sigma points on the joint distribution $bel(\psi_{t-1}, \phi_{t-1})$ instead of using standard Unscented Transform points: 
\begin{equation}
     \chi^{[i]}_{t-1} = \mu_{t-1} + \epsilon^{[i]} \sqrt{\Sigma_{t-1}} 
     \label{eq:mc_sampling}
\end{equation}
where, $i \in 1..C$ and $\chi_{t-1} = [\chi_{\psi_{t-1}}, \chi_{\phi_{t-1}}] \in \mathbb{R}^{Cx8}$ with  an associated weight $w_t^{[i]} = 1/C$ and $\epsilon^{[i]} \sim \mathcal{N}(0, 1)$. We set $C=100$ for all our experiments. The sigma points are filtered based on whether they satisfy the physical constraints and passed through the data-driven models. However, the invalid sigma points are also retained and reintroduced to preserve the uncertainty of the distribution. This is visually illustrated and explained in Appendix - Fig. \ref{fig:mc_sampling}.

 \begin{figure}[t!]
    \centering
    \includegraphics[width = \columnwidth, height = 4.8
    cm]{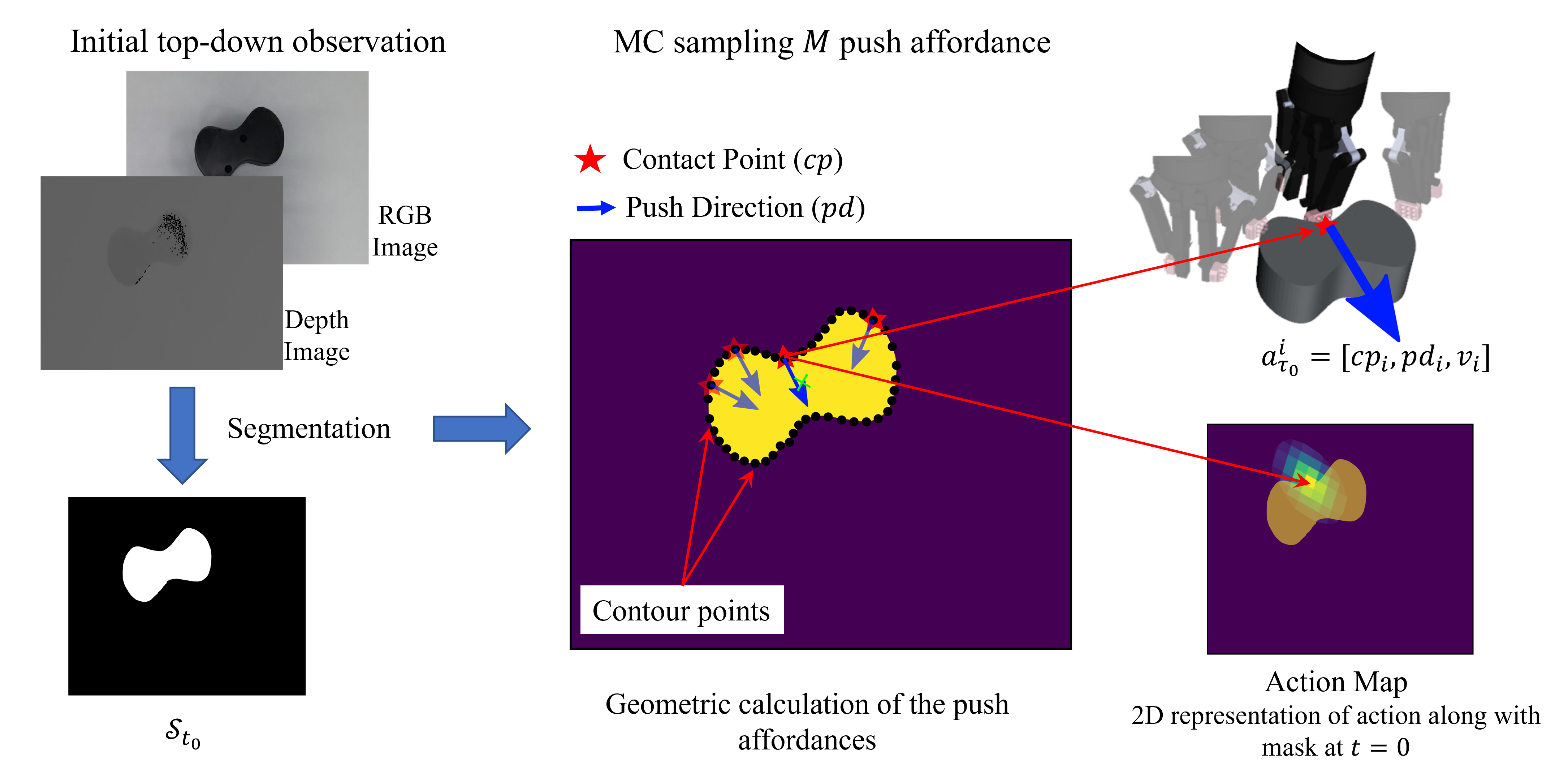}
    \caption{Monte-Carlo (MC) action sampling and action maps illustration}
    \label{fig:action_sampling}
\end{figure}

\textit{Shape-Action Encoder}: During the pushing, it is important to take into account the local geometry of the object and the action. Few previous works \cite{planar_push_data, alina_1} have shown that such an approach improves the action effect prediction. We encode the action along with the local geometry of the object at the point of contact to improve action effect predictions via the \textit{ShapeActionEncoderNet}. 

\textit{ShapeActionEncoderNet: } 
This comprises 3-layer CNN layers followed by 2 layers of feed-forward neural network. For each sigma point sampled from the current belief, an expected \textit{segmentation mask} $\hat{\mathcal{S}}_t$ is generated by transforming the initial segmentation $\mathcal{S}_0$ based on the pose information $\chi_{\psi_{t-1}}$. This represents the current geometry of the object at the point of action. Next, a 2D representation of the action - \textit{action map} $\mathcal{M}_t$ is generated. This is done via representing a 2D Gaussian distribution based on the action affordance $a_t = (cp, pd, v)$ in the image frame. Further details of the action map can be found in the appendix, and visually depicted in Fig.\ref{fig:action_sampling}. By this approach, we avoid generating complex object shape predictions for intricate visual perspective. 


\textit{ProcessNet:}
\label{sec:process}
The data-driven process model for predicting the change in \textit{pose} of the object is approximated via 3 layer feed-forward neural network given prior joint sigma points and the shape-action encoding $enc_t$.  In addition, we also employ the learnt heteroscedastic process noise model. 
\vspace{-5pt}
\begin{flalign}
    enc_{a_t} & \longleftarrow ShapeActionEncoderNet([\hat{\mathcal{S}}_t, \mathcal{M}_t]) \\
    \overline{\chi}_{\psi_t}, Q_t &\longleftarrow ProcessNet(\chi_{t-1}, enc_{a_t}) \\
    \overline{\chi}_{\phi_t} &= {\chi}_{\phi_{t-1}}
     \label{eq:process_model}
\end{flalign} 
where, $Q_t \in \mathbb{R}^{Cx3}$ is the heteroscedastic diagonal covariance noise for time-varying pose. The predicted next step sigma points $\overline{\chi}_t$, along with the process noise $Q_t$ are utilized to compute the expected Gaussian belief $\overline{bel}(\psi_t, \phi_t)$ as 
\begin{flalign}
    \overline{\chi}^{[i]}_{\psi_t} &= \overline{\chi}^{[i]}_{\psi_t} + \epsilon^{[i]} \sqrt{Q_t^{[i]}} \\
    \overline{\mu}_t &= \sum_{i=0}^{C}w_t^{[i]} \overline{\chi}_t \\
	\overline{\Sigma}_t &= \sum_{i=0}^{C}w_t^{[i]}(\overline{\chi}^{[i]}_t-\overline{\mu}_t)(\overline{\chi}^{[i]}_t-\overline{\mu}_t)^T
     \label{eq:predicted_belief}
\end{flalign} 
where, $i \in 1.. C$ and $\overline\chi_{t} = [\overline\chi_{\psi_{t}}, \overline\chi_{\phi_{t}}]$

\subsubsection{Update Step}
We recompute the constrained Monte-Carlo sigma point $\overline{\chi}^{'}_{\phi_t}$ sampling on the predicted belief $\overline{bel}(\psi_t, \phi_t)$ to incorporate the process noise. The dual filter employs a separate update of parameter belief similar to the parameter update presented in \cite{liu_west} and the conditional pose belief update based on the UKF update \cite{prob_rob_book}.  For updating the joint belief, we require an observation model to predict observation sigma points $\overline{\mathcal{Z}}_t$ which has to account for both visual and tactile observations. To reduce the complexity of predicting raw RGB-D images, we split into the observation model two components, tactile and visual models. A \textit{VisNet} network acts as a synthetic sensor generating the current noisy 2D pose information  $x, y, \theta$ from the current RGB-D images at each time. The \textit{VisNet} comprises of first 10 layers of VGG-19 \cite{simonyan2014very} pretrained on ImageNet followed by 3 layers of feed-forward network. For the tactile counterpart, a 4 layers of feed-forward network  \textit{TacNet} is utilised to predict the contact force information. In addition, a two-layer network \textit{ObsNoiseNet} is also utilised to generate the heteroscedastic and diagonal observation noise. 

\begin{flalign}
     \overline{\mathcal{Z}}^{V}_t &= \overline{\chi}^{'}_{{\psi}_t}
     \overline{\mathcal{Z}}^{T}_t \longleftarrow  TacNet(\overline{\chi}^{'}_t, enc_{a_t})\\
     z^V_t & \longleftarrow VisNet(o^V_t), z^T_t = o^T_t \\
     R_t &\longleftarrow ObsNoiseNet(z^V_t, z^T_t)
     \label{eq:observation_model}
\end{flalign}

\subsubsection*{Parameter Update}
We update the weights based on the likelihood of the observation sigma points $\overline{\mathcal{Z}}_t = [\overline{\mathcal{Z}}^{T}_t, \overline{\mathcal{Z}}^{V}_t] $ in the observation distribution $\sim \mathcal{N}(.|z_t, Q_t)$
\begin{flalign}
    w_t^{[j]} &=  w_t^{[j]}e^{(-\frac{1}{2}(\overline{\mathcal{Z}}^{[j]}_t - z_t) R^{-1}(\overline{\mathcal{Z}}^{[j]}_t - z_t)^T)}
     \label{eq:parameter_update}
\end{flalign}
where $j \in 1.. C$. The updated parameter belief $bel(\phi_t)$ is recomputed via a Gaussian Smooth Kernel \cite{liu_west} method after normalizing the updated weights.
\begin{flalign}
    \mu_{\phi_t} &= \sum_{i=0}^{C}w_t^{[i]} \overline{\chi}^{'}_{\phi_t};  \quad m^{[i]}_{\phi_t} = a\overline{\chi}^{'}_{{\phi}_t}+(1-a)\mu_{\phi_t} \\
	\Sigma_{\phi_t} &= h^2\sum_{i=0}^{C}w_t^{[i]}m^{[i]}_{\phi_t}-\mu_{\phi_t}
\end{flalign}
where $a$ and $h=\sqrt{1-a^2}$ are shrinkage values of the kernels that are user-defined and set to 0.01, and $m$ are the kernel locations.

\subsubsection*{Pose Update}
We make use of the dependence of the pose on the parameters to compute the conditional pose distribution $bel(\psi_t | \phi_t) \sim \mathcal{N}(\psi_t|\mu_{\psi_t|\phi_t}, \Sigma_{\psi_t|\phi_t})$ using Multivariate Gaussian Theorem \cite{mult_gauss}. 
\begin{flalign}
    \mu_{\psi_t|\phi_t} &= \psi_t + \Sigma_{\psi_t\phi_t}\Sigma^{-1}_{\phi}(\phi_t - \mu_{\phi_t})\\
    \Sigma_{\psi_t|\phi_t} &= \Sigma_{\psi_t} -  \Sigma_{\psi_t\phi_t}\Sigma^{-1}_{\phi_t}\Sigma_{\phi_t\psi_t}
    \label{eq:cond_pose_distribution}
\end{flalign}

For conditional pose update, standard Unscented Kalman Filter (UKF) is employed on the predicted conditional pose distribution $\overline{bel}(\psi_t | \phi_t = \mu_{\phi_t})$ using Eq.\ref{eq:cond_pose_distribution}. The $\mu_{\phi_t}$ of the updated parameter belief is utilised with predicted pose sigma points $\overline{\chi}^{UT}_{\psi_t}$ to obtain the predicted observation sigma points $\overline{\mathcal{Z}}'_t$. The UKF update equations are skipped for brevity. After the conditional pose update, the posterior joint is computed as:

\begin{equation}
     bel(\psi_t, \phi_t) = bel(\psi_t|\phi_t)bel(\phi_t)
     \label{eq:belief_joint_update}
 \end{equation}
Note, the cross-covariance matrices $\Sigma_{\psi_t\phi_t}, \Sigma_{\phi_t\psi_t}$ are not updated through the dual update step and are kept constant.

\subsection{Active Non-prehensile Pushing Actions}
The push action affordance is given by the tuple $a_t = (cp, pd, v)$. Possible \textit{contact point's} $cp$ and normal angle $cn$ at the contact point is geometrically computed from the initial 2D segmentation $\mathcal{S}_0$ based on our previous work in ~\cite{murali_interactive} and illustrated in Fig.\ref{fig:action_sampling}.

\textit{Monte-Carlo Sampling of push affordance}:
We generate $M$ push affordances, $a^{[i]}_t$, $i \in 1.. M$,  from the possible points of contact points and contact normal by sampling a contact point and generating the $pd^{[i]} = cn^{[i]} + \delta; \delta \sim R(-5, 5)$ (deg). The velocity $v$ is fixed for all cases keeping in mind the quasi-static assumption. 

\textit{$N$-step Information Gain:}
To make the framework more sample efficient for real robot scenarios, we employ active action selection by formulating an $N$-step information gain criteria \cite{fep} under the filtering setting. We recursively use the prediction step of the dual differentiable filter without the update step to compute the expected Information Gain for both model learning and object parameter inference for each sampled non-prehensile pushing action $\pi^{[i]}$ = $a^i_{\tau_0:\tau_N}$ over $N-$step in future $\tau = \tau_0.. \tau_N$ 
\begin{align}
     IG_{N}(\pi^{[i]}) &\approx  - \mathbb{E}_{p(\psi_{\tau_N}, \phi_{\tau_N}|\pi^{[i]})}[ln(\overline{bel}^{[i]}(\psi_{\tau_N}, \phi_{\tau_N}) \nonumber \\
     &- ln(\overline{bel}^{[i]}(\psi_{\tau_0} , \phi_{\tau_0})]
     \label{eq:IG_N_step}
\end{align}
where, $\overline{bel}^{[i]}(\psi_{\tau_N}, \phi_{\tau_N})$ is the hypothetical predictive joint distribution after $N$-step by taking action $\pi^{[i]}$ without taking account the actual observation. For our case, the expectation is computed as KL-Divergence form for which the closed form solution exists for Multivariate Gaussian distributions \cite{kl_gauss}. 
\vspace{-3pt}
\begin{flalign}
     IG_{N}(\pi^{[i]}) \approx D_{KL}&[\mathcal{N}^{[i]}(\psi_{\tau_N}, \phi_{\tau_N} | \overline{\mu}_{\tau_N}, \overline{\Sigma}_{\tau_N})
     || \mathcal{N}^{[i]}(\psi_{\tau_0} , \phi_{\tau_0} | \overline{\mu}_{\tau_0}, \overline{\Sigma}_{\tau_0})] \nonumber \\
     \pi^* &= \argmax_{\pi^i} IG_N(\pi^{[i]}) 
     \label{eq:active_action}
\end{flalign}

\section{Experiments}
\label{sec:results}

In this section, we explain the experiment setup and the results obtained from the proposed method hereby referred to as VT-ADDF (Visuo-Tactile Active Dual Diffentiable Filter). The closest state-of-the-art work to ours which dealt with the estimation of object parameters using robotic pushing was that of \cite{nikos_base_work} and have taken it as the baseline. The baseline work utilised feature extraction using object pose, actions, and contact force information and a Multi-Output Regression Random Forest for data-driven regression modelling. We re-implemented the baseline approach to the best of our capability and validated the published results on the MIT Push Dataset. 
\begin{figure}[t!]
    \centering
    \includegraphics[width = \columnwidth, height = 4.2
    cm]{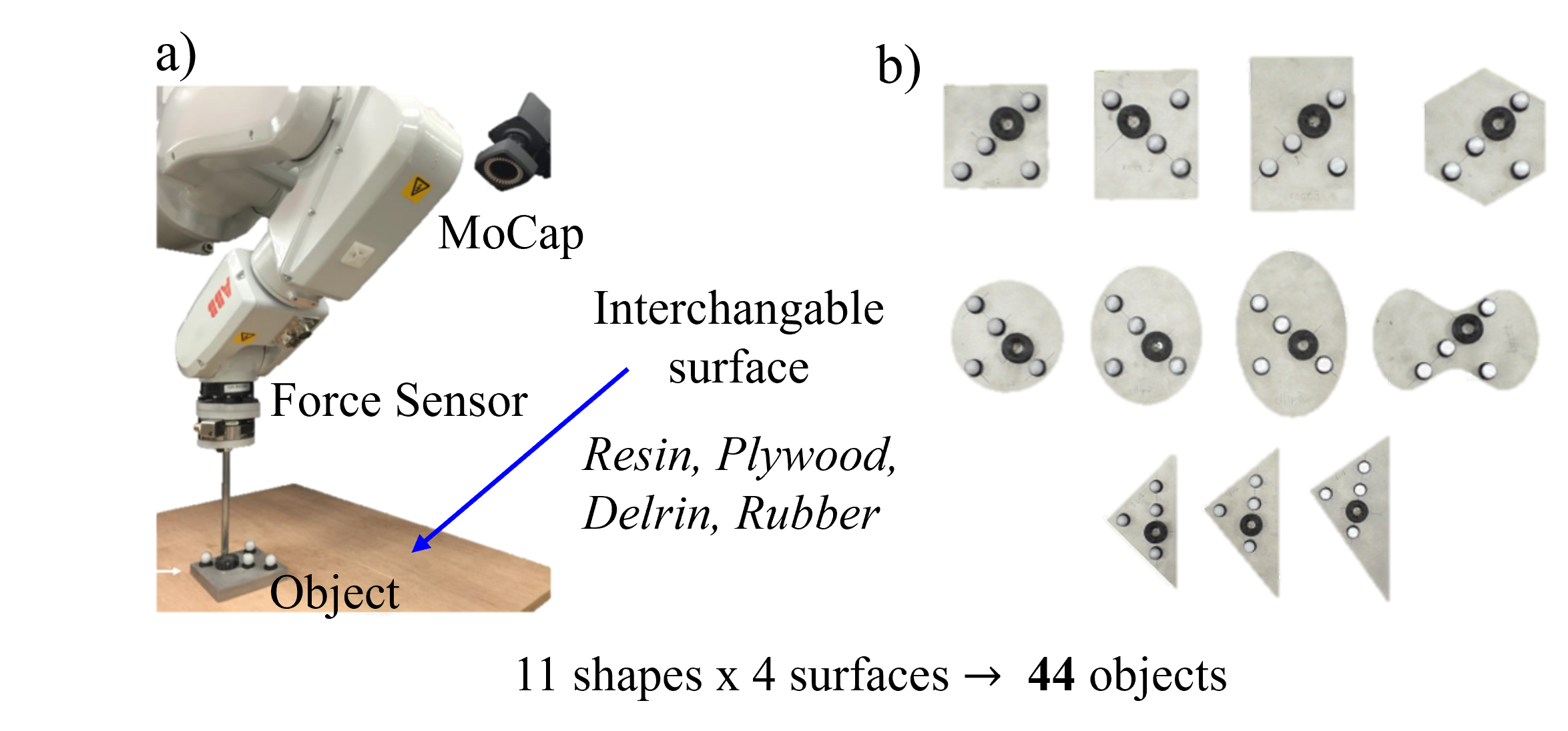}
    \caption{MIT Push Dataset Setup \cite{mcubemit:online}. Part a) presents the data collection setup. Part b) presents the various objects in the dataset}
    \label{fig:mit_push_database}
\end{figure}

In addition, we performed extensive ablation studies 1) exploring the efficiency of the active approach vs random and uniform actions for learning and inference, 2) employing only vision for parameter estimation under the dual filtering setup (termed at V-DDF Visual-Dual Differentiable Filtering). For this, the \textit{TacNet} was removed and the observations were reduced to only RGB-D. The rest of the framework and dual filtering setup with the active actions remained the same. 
3) Study of dual filtering approach compared to joint filtering (termed as VT-JDF). In this, instead of performing separate parameter and pose updates, only a single UKF update equation was used \cite{alina_2_main}.

\subsection{Experimental Setups}
We tested our approach and compared the baseline on 3 experimental setups. 

\subsubsection{Dataset - MIT Push Dataset}
We utilised the MIT Push dataset, a state-of-the-art robotic pushing dataset \cite{mit_push_d} with 44 different objects as shown in Fig.\ref{fig:mit_push_database} The dataset contains tactile, synthetic RGB-D, pose as well as parameter information. The objects were of 11 different shapes with varying mass, inertia, and 4 different surfaces - Abs, Delrin, Plywood, and Rubber-sheet were present. The center of mass of each object was slightly varied w.r.t its geometric center. The dataset has almost 10,000 pushes for each object, however, we selected a partial subset of pushes with no acceleration and velocity of 30 mm/s in a total of 3750 pushes. As this was a pre-recorded dataset, we only present estimation with uniform actions rather than active actions. We selected this setup to validate the baseline results as well as showcase that our proposed visuo-tactile dual differentiable filter can be utilized for different robotic environments.  

\subsubsection{Simulation Setup - \textit{Sim Robotac}}
\begin{figure}[t!]
    \centering
    \includegraphics[width = \columnwidth, height = 4 
    cm]{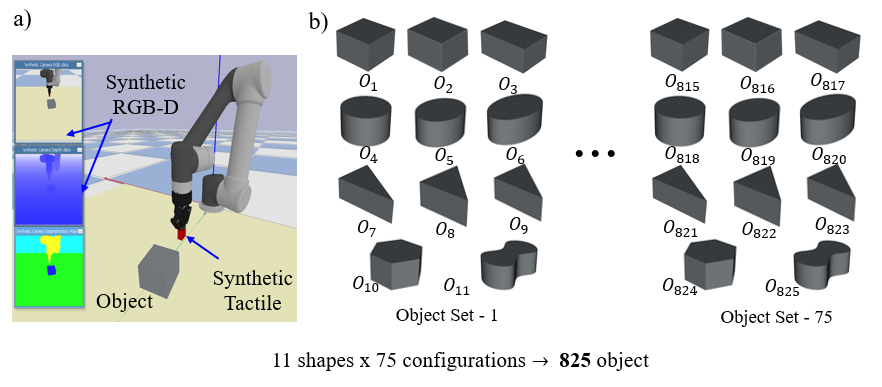}
    \caption{Simulation Setup-\textit{Sim Robotac}. Part a) presents the PyBullet scene of the setup. Part b) presents the object set (825) objects used for the experiments.}
    \label{fig:sim_robotac_setup}
\end{figure}

We designed a simulation setup in PyBullet \cite{pybullet} to evaluate extensively our proposed approach as shown in Fig.\ref{fig:sim_robotac_setup}. In addition, it is possible to have a large set of objects with variations in physical parameters in the simulation which is often difficult real robotic setup. The setup comprised a simulated Robotiq gripper mounted on UR5 with a simulated tactile sensor attached to one of the finger pads. A synthetic RGB-D sensor was placed on top of the pushing area to simulate a visual sensor. In the simulation setup, 825 different objects were designed based on the parameters as presented in Table \ref{tab:sim_parameters}. We utilised the simulation setup to perform extensive ablation studies, the results of which are presented in the following section.

\begin{table}[t!]
\centering
\caption{Parameter range for simulation setup}
\label{tab:sim_parameters}
\begin{tabular}{|l|l|}
\hline
\multicolumn{1}{|c|}{\textbf{Property}} & \multicolumn{1}{c|}{\textbf{Range of values}} \\ \hline
$Mass$ (kg)                               & {[}0.2, 0.5, 0.8, 1.2{]}                           \\ \hline
$\mu$                                      & {[}0.35, 0.5, 0.7{]}                          \\ \hline
$COM_x$ (m)                              & {[}-0.02, -0.015, 0, 0.015, 0.02{]}           \\ \hline
$COM_y$ (m)                              & {[}-0.02, -0.01, 0, 0.02, 0.03{]}             \\ \hline
$I_z$ (g.m2)                             & {[}0.5, 0.9, 1.15, 1.5{]}                     \\ \hline
$Shapes$                                  & 11 (Fig. \ref{fig:sim_robotac_setup} )                                    \\ \hline
\end{tabular}
\end{table}
\vspace{-5pt}
\subsubsection{Robotic Setup}
\begin{figure}[t!]
    \centering
    \includegraphics[width = \columnwidth, height = 5
    cm]{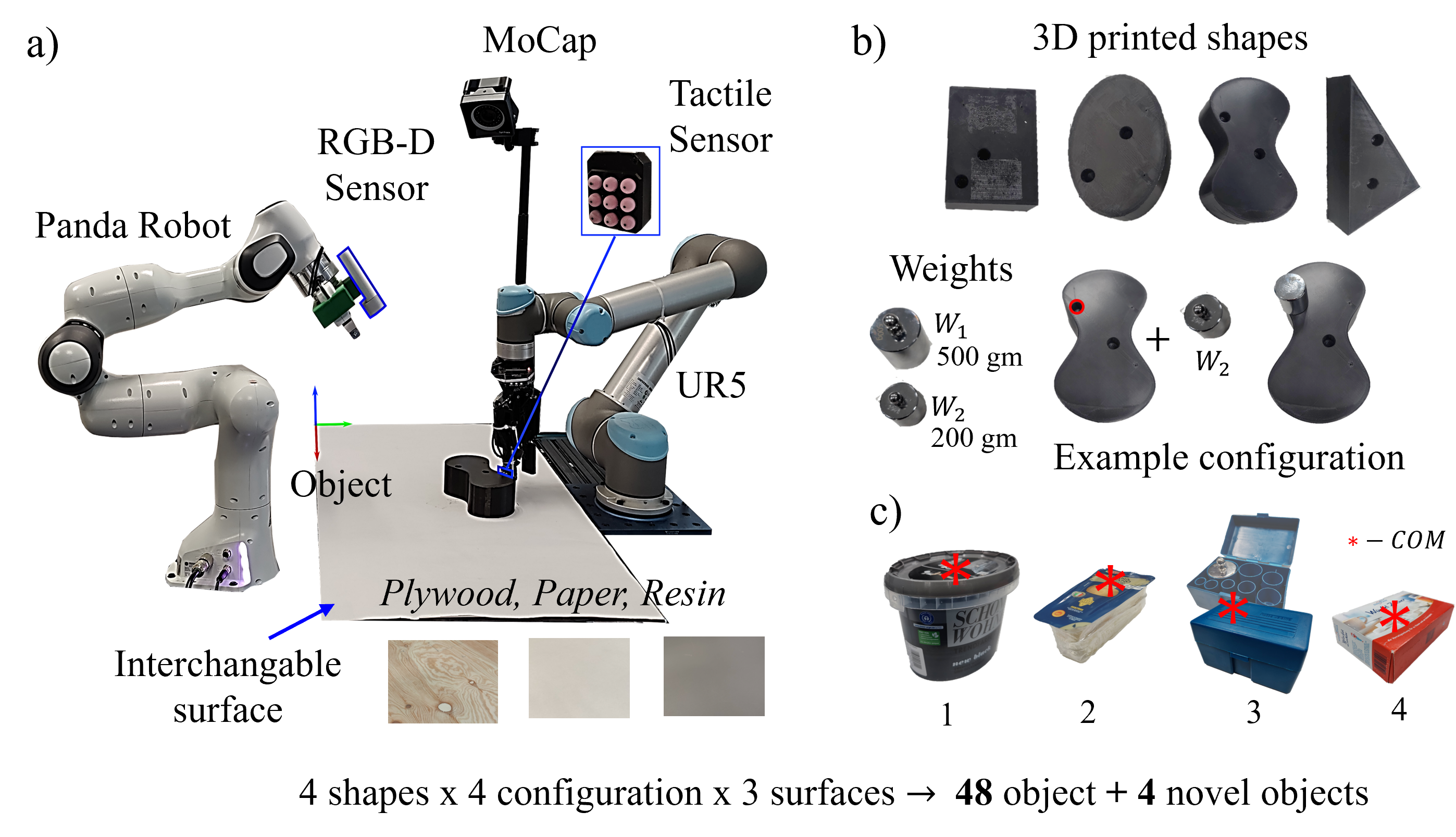}
    \caption{Robotic Setup - Real Robotac. Part a) presents the robotic setup. Part b) presents an overview of the configurable objects. Part c) presents the novel objects selected to test}
    \label{fig:real_robotac_setup}
\end{figure}

The robotic setup consists of Universal Robots (UR5) augmented with Robotiq two-finger Gripper and a Panda robotic manipulator as shown in Figure \ref{fig:real_robotac_setup}. Tactile sensor \cite{contactile} is attached to the outer surface of the finger of the gripper pads of the Robotiq Gripper and an Azure DK RGB-D camera is rigidly attached to the Panda Gripper. The maximum allowed speed for the UR5 was 25 mm/s for safety constraints. The ground truth values of the pose were collected using the motion capture system - Optitrack \cite{OptiTrac66:online}, whereas the ground truth values of the object parameters were computed from a CAD model of the objects. To obtain real objects with varying parameters, we designed configurable objects by 3D printing 4 shapes and adding additional weights at a precise location in the objects, changing their mass, center of mass and inertia value. In addition, we utilised 3 different frictional surfaces - plywood, paper and resin sheet, to vary the relative friction coefficient between the object and the pushing surfaces. In total, we had 48 different objects after all possible configuration. In addition, we have 4 novel daily objects as shown in Figure.\ref{fig:real_robotac_setup} (c) which were not used in training and were kept only for testing. The objects had contrasting parameters (high mass of paint box (Object 1), high friction of sugar cube box (Object 4) and the plywood surface, as well as shifted COM in cheese (Object 2) and the weight box (Object 3). 

\subsection{Experimental Results}
\subsubsection{Learning - DDF Training}
For training the networks used in the differentiable filter, we utilised a weighted combination of negative log-likelihood  loss $\mathcal{L_{NLL}}$ of the ground pose and parameter w.r.t to the  belief and mean squared error loss $\mathcal{L_{MSE}}$ of contact forces and synthetic pose. Iterative training was done using Adam optimizer till the loss converged. 

\begin{figure}[t!]
    \centering
    \includegraphics[width = \columnwidth, height = 4
    cm]{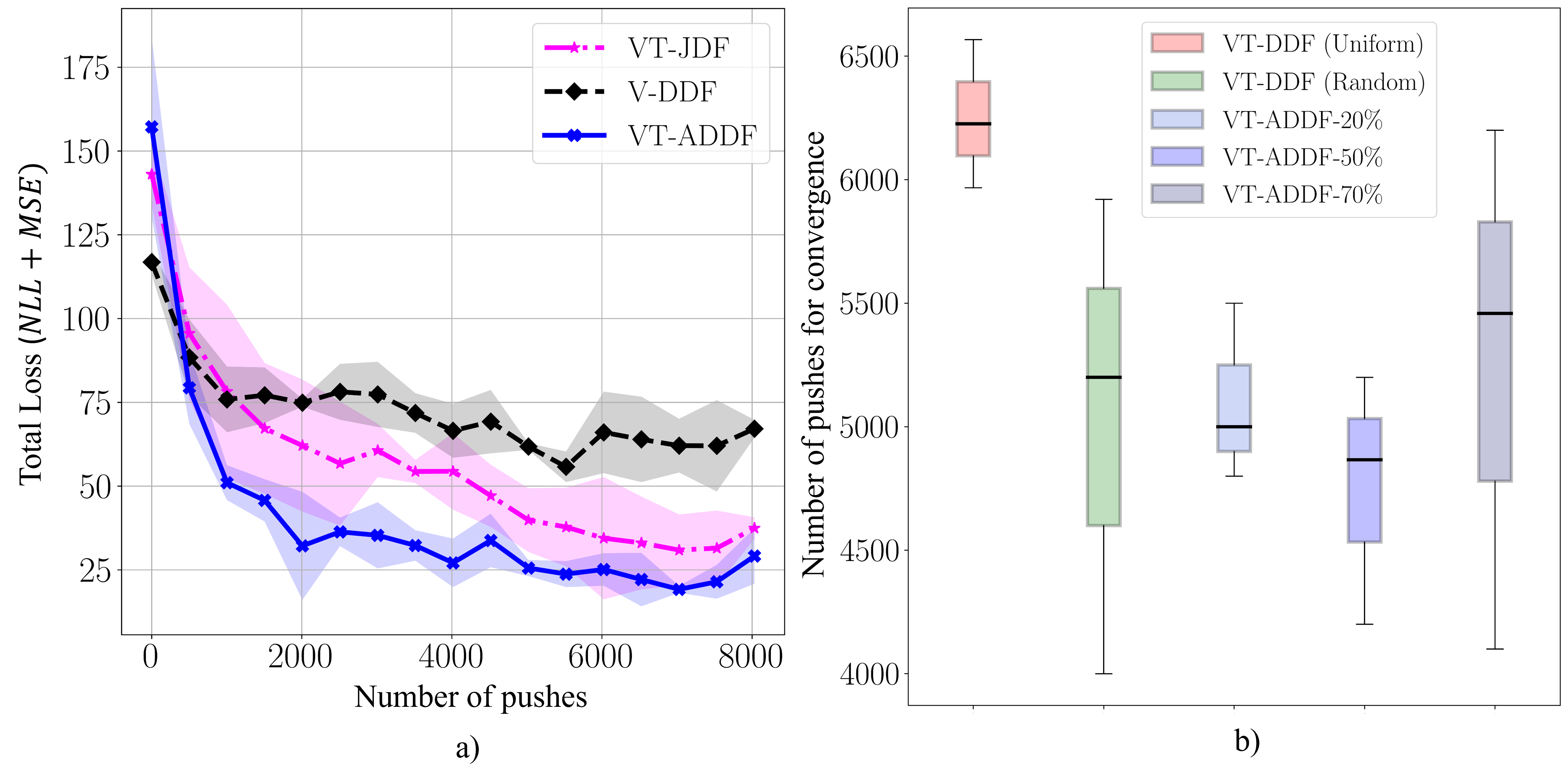}
    \caption{Learning results of the ablation studies in Sim-Robotac setup. Part (a) presents comparitive performance learning stability of VT-ADDF vs V-DDF vs VT-JDF. Part (b) presents the learning efficiency of different push action selection methods - Uniform, Active, Random}
    \label{fig:Learning_plot}
\end{figure}

For MIT Push Dataset, the time horizon was $t_H=10s$ with a sampling rate of 10Hz. We split the 3750 trajectories, into 90\% for training and 10\% for inference. For the Sim-Robotac setup, the time horizon was $t_H=15s$ with a sampling rate of 10Hz. 90\% of 825 objects were utilised for training and 10\% for testing, cross-validated 5 times. We performed an ablation study utilising a uniform, random and active approach of taking the action from the set of $M$-push affordances for training the filter. In addition, we also explored how much $N$-step lookahead is suitable. We chose $N$ as 20\%(=3 secs), 50\% (=7.5 secs), and 70 (=10.5)\% of the time horizon as future look-ahead steps for ablation study on active actions. We present the results of the ablation study in Fig. \ref{fig:Learning_plot} (b), for learning efficiency. 
In addition, we also present the validation loss plots to highlight the stability and learning performance of the proposed approach (VT-ADDF) compared to using only vision (V-DDF) and utilizing a joint differentiable filter (VT-JDF) in Figure. \ref{fig:Learning_plot} (a). 
For the Real-Robotac setup, the time horizon was $t_H=10s$, with a sampling rate of 5 Hz and an active approach with 50\% $N$-step lookahead (=5secs) selected for training the dual differentiable filters. 90\% of the 48 3D printed configurable objects were utilised for training and 10\% for testing.

\subsection{Parameter Inference}
\begin{figure*}[t!]
    \centering
    \includegraphics[width = \textwidth, height = 6.8
    cm]{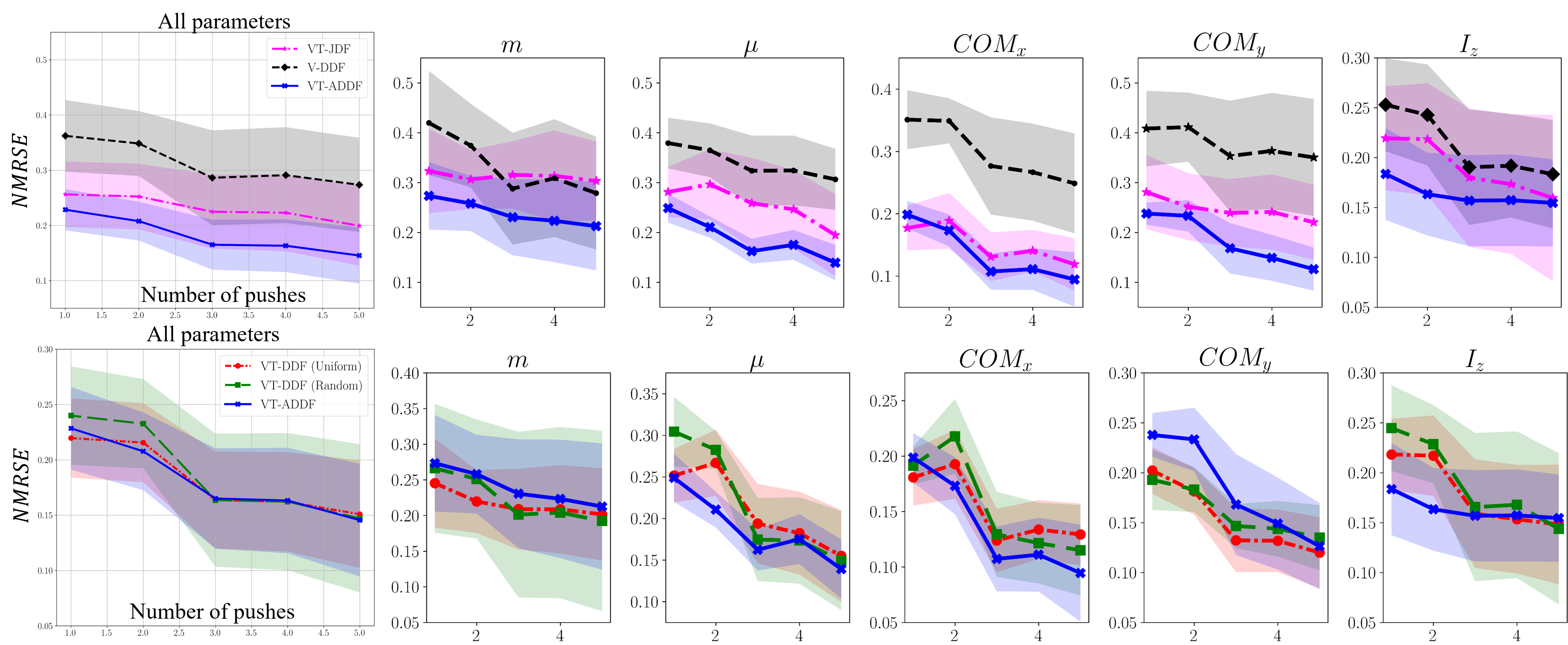}
    \caption{Inference result during the filtering step presented after each push action. }
    \label{fig:results_inference_ablation}
\end{figure*}

For parameter inference of unknown (test) objects, we executed multiple push actions. At the end of each, the posterior belief of object parameters was utilised to initialize the belief for the next push. We present the results of the parameters $m, \mu, CoM_x, CoM_y, Iz$ inference for the ablation study in Sim-Robotac setup in Table.\ref{tab:sim_ablation}. For parameter inference, the $N$-step lookahead was selected as 50\% of time horizon $t_H$ for both Sim-Robotac and Real-Robotac setups. In addition, Fig.\ref{fig:results_inference_ablation} presents a closer look at the filtering action of the different ablation approaches and parameter inference convergence in the Sim-RoboTac setup. Further, the comparative parameter estimation results of the proposed approach (VT-ADDF) compared to the baseline work of \cite{nikos_base_work} are presented in  Table.\ref{tab:basel_linke_results}. We present separate estimation results for the novel objects which were not utilised for training, to evaluate the generalisation of the proposed approach compared to the baseline. As the range of values of different parameters like mass, friction co-efficient and inertia are different and the values often close to 0 (for the center of mass) we employ a normalised root mean square $NRMSE$ \cite{nikos_base_work} as metric to evaluate the performance over different parameters which is the root mean squared error divided by the range of values $(\psi_{max}-\psi_{min})$ of each parameters in the setups. Lower value, signifies better estimation.

\begin{table*}[ht!]
\centering
\caption{Inference result of $NRMSE$ values of different parameters in the ablation study in Sim-Robotac setup}
\label{tab:sim_ablation}
\begin{tabular}{|l|l|l|l|l|l|l|}
\hline
                 & $mass$                & $\mu$                  & $com_x$              & $com_y$              & $I_z$                & Overall             \\ \hline
VT-JDF           & 0.33 $\pm$ 0.08          & 0.19 $\pm$ 0.08          & 0.12 $\pm$ 0.04          & 0.22 $\pm$ 0.08          & 0.16 $\pm$ 0.08          & 0.20 $\pm$ 0.07          \\ \hline
V-DDF            & 0.28$\pm$0.11          & 0.31$\pm$0.06          & 0.25$\pm$0.08          & 0.35$\pm$0.12          & 0.18$\pm$0.05          & 0.27$\pm$0.09          \\ \hline
VT-ADDF          & \textbf{0.21}$\pm$\textbf{0.09} & \textbf{0.14}$\pm$\textbf{0.04} & \textbf{0.09}$\pm$\textbf{0.04} & \textbf{0.13}$\pm$\textbf{0.04} & \textbf{0.15}$\pm$\textbf{0.03} & \textbf{0.14}$\pm$\textbf{0.05} \\ \hline
VT-DDF (Uniform) & 0.20$\pm$0.06          & 0.16$\pm$0.06          & 0.13$\pm$0.03          & 0.12$\pm$0.03          & 0.12$\pm$0.06          & 0.15$\pm$0.05          \\ \hline
VT-DDF (Random)  & 0.19$\pm$0.13          & 0.15$\pm$0.06          & 0.11$\pm$0.04          & 0.13$\pm$0.03          & 0.14$\pm$0.08          & 0.14$\pm$0.07          \\ \hline
\end{tabular}
\end{table*}

\begin{table*}[ht!]
\centering
\caption{Parameter Inference result of $NRMSE$ value for proposed approach VT-ADDF compared to baseline work of \cite{nikos_base_work} in various setups}
\label{tab:basel_linke_results}
\resizebox{\textwidth}{!}{%
\begin{tabular}{|l|ll|ll|ll|ll|ll|}
\hline
& \multicolumn{2}{c|}{$mass$}  & \multicolumn{2}{c|}{$mu$}    & \multicolumn{2}{c|}{$com_x$}  & \multicolumn{2}{c|}{$com_y$}  & \multicolumn{2}{c|}{$I_z$} \\ \hline
\textit{Experimental Setup}    & \multicolumn{1}{l|}{Baseline} & VT-ADDF & \multicolumn{1}{l|}{Baseline} & VT-ADDF             & \multicolumn{1}{l|}{Baseline}  & VT-ADDF & \multicolumn{1}{l|}{Baseline} & VT-ADDF & \multicolumn{1}{l|}{Baseline} & VT-ADDF \\ \hline
MIT Push Dataset  & \multicolumn{1}{l|}{0.11$\pm$0.1} & 0.19$\pm$0.02 & \multicolumn{1}{l|}{0.18$\pm$0.04} & \textbf{0.17}$\pm$\textbf{0.02} & \multicolumn{1}{l|}{0.13$\pm$0.06} & \textbf{0.10}$\pm$\textbf{0.04} & \multicolumn{1}{l|}{0.12$\pm$0.09}  & \textbf{0.09}$\pm$\textbf{0.07} & \multicolumn{1}{l|}{0.17$\pm$0.02} & \textbf{0.16}$\pm$\textbf{0.01}  \\ \hline
Sim Robotac  & \multicolumn{1}{l|}{0.14$\pm$0.06} & 0.21$\pm$0.09 & \multicolumn{1}{l|}{0.16$\pm$0.06} & \textbf{0.14}$\pm$\textbf{0.04} & \multicolumn{1}{l|}{0.18$\pm$0.12} & \textbf{0.09}$\pm$\textbf{0.04} & \multicolumn{1}{l|}{0.14$\pm$0.15}  & \textbf{0.13}$\pm$\textbf{0.04} & \multicolumn{1}{l|}{0.20$\pm$0.11} & \textbf{0.15}$\pm$\textbf{0.03} \\ \hline
Real Robotac (RR)  & \multicolumn{1}{l|}{0.25$\pm$0.12}  & \textbf{0.22}$\pm$\textbf{0.09} & \multicolumn{1}{l|}{0.14$\pm$0.03}  & 0.19$\pm$0.06          & \multicolumn{1}{l|}{0.12$\pm$0.08} & \textbf{0.10}$\pm$\textbf{0.01} & \multicolumn{1}{l|}{0.20$\pm$0.1} & \textbf{0.11}$\pm$\textbf{0.07} & \multicolumn{1}{l|}{0.16$\pm$0.05} & \textbf{0.09}$\pm$\textbf{0.01} \\ \hline
RR Novel Objects & \multicolumn{1}{l|}{0.29$\pm$0.09}  & \textbf{0.20}$\pm$\textbf{0.05} & \multicolumn{1}{l|}{0.21$\pm$0.03}  & \textbf{0.18}$\pm$\textbf{0.06} & \multicolumn{1}{l|}{0.17$\pm$0.09} & \textbf{0.11}$\pm$\textbf{0.05} & \multicolumn{1}{l|}{0.22$\pm$0.05} & \textbf{0.12}$\pm$\textbf{0.05} & \multicolumn{1}{l|}{0.22$\pm$0.10} & \textbf{0.15}$\pm$\textbf{0.08} \\ \hline
\end{tabular}
}%
\end{table*}

\subsection{Discussion}
In this work, we proposed a novel visuo-tactile based active object parameter inference with a dual differentiable filter.

The results of the ablation study for learning show that active actions significantly improve the sample efficiency by around 20\% (push actions) compared to uniform actions and has lower variance than random action selection as presented in Fig.\ref{fig:Learning_plot}(b). Moreover, it is shown that the dual filtering approach has more stable learning than the joint filter as presented in Fig.\ref{fig:Learning_plot}(a). This demonstrates the efficacy of our proposed novel dual differentiable filtering approach compared to a joint differentiable filter method. Furthermore, our experimental results show that by using only vision, the network fails to reduce the loss and tends to overfit. This is expected, as the parameter estimation is difficult only via vision and has high error rates, leading to higher loss values. 

The obtained results from the parameter inference show that our proposed approach performs consistently on different experimental setups - MIT Push Dataset, Sim-Robotac and Real-Robotac, compared to baseline work of \cite{nikos_base_work}, which fails to generalise for novel objects in real robotic setup. Moreover, the limitation of providing ground truth pose information in the baseline approach is elevated by our proposed ADDF approach during the inference step. Furthermore, the ablation study shows that active actions have a better estimation of parameters compared to uniform and with lower variance than random actions with the same number of push actions. Compared to using only vision, the visuo-tactile dual differentiable approach performs much better, especially in parameters like the center of mass prediction, as well as, is more stable and accurate than the joint filtering approach. Through the different setups, we also show that the proposed visual tactile-based dual differentiable filter for parameter inference is agnostic to robotic setups as long as sufficient visual and tactile information is present. 

From the limitations, our proposed ADDF requires separate training of the \textit{VisNet} with $MSE$ loss to obtain pose information for the novel objects which are visually quite different from the training set. Instead of using RGB-D as visual observations and a 2D pose estimation network, it will be viable to use point clouds and avoid the requirement of using pose estimation altogether or use recent one-shot pose-estimation approaches. In addition, it will be interesting to avoid the requirement of having ground truth states and parameter values during training as well as develop a framework which can discover physical object representations.

\section{Conclusion}
In this work, we addressed the problem of estimating the proprieties of rigid objects using vision and tactile observations solely via non-prehensile pushing. The proposed approach first learns an object interaction model using known objects, which is utilized for inference of novel objects under differentiable filter settings. We present a novel formulation of active action selection with the differentiable filter as one of the key contributions. The generalizable capability of the framework makes it viable for real robotic applications and opens the possibility to explore the approach for other interaction techniques for object parameter estimation like grasping. 

\section*{Acknowledgment}
We would like to thank Dr. Alina Kloss for her useful insights and open sourced implementation of the differentiable filters. In addition, we would like to thank Prajval Kumar Murali, Iman Nematollahi, and Dr. Xiaoxiao Cheng for their constructive reviews.  

\bibliography{biblio}
\bibliographystyle{IEEEtran}

\section*{Appendix}
\textbf{Constrained Monte Carlo Sampling}
\begin{figure}[h]
    \centering
    \includegraphics[width = \columnwidth, height = 70
    mm]{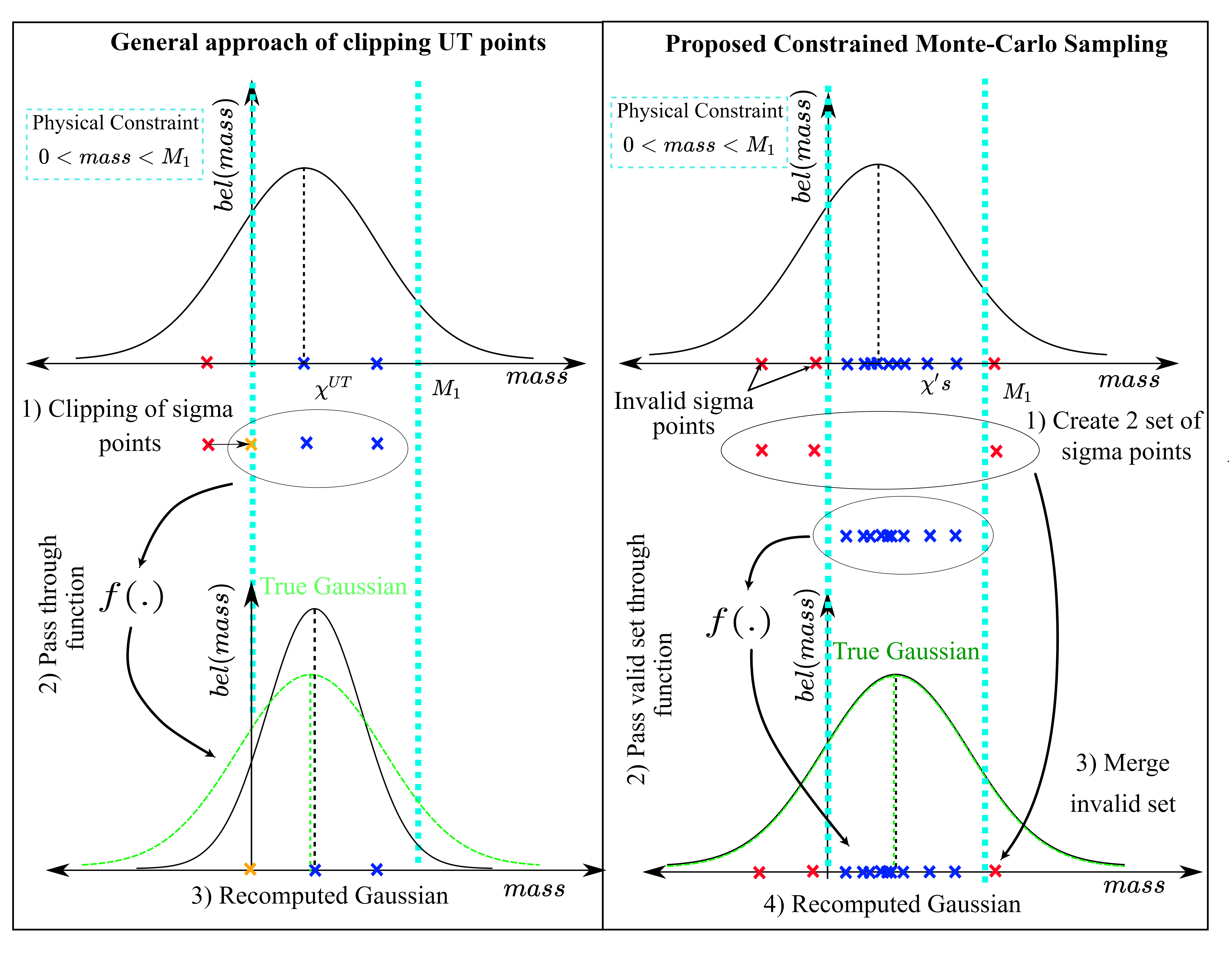}
    \caption{Constrained Monte Carlo sampling}
    \label{fig:mc_sampling}
\end{figure}

\textbf{Action Maps}
A 2D Gaussian can be generated by the following equations, along the push direction $pd$ and centered at the contact point in the image frame ($cp_{if}$).
\begin{align*}
px &= \begin{pmatrix}
pixel_x \\
pixel_y
\end{pmatrix},  K = \begin{pmatrix}
\frac{cos^2{(pd)}}{2v^2}+\frac{sin^2(pd)}{2} & \frac{sin{(2pd)}}{4v^2}-\frac{sin(2pd)}{4} \\
\frac{sin{(2pd)}}{4v^2}-\frac{sin(2pd)}{4} & \frac{sin^2{(pd)}}{2v^2}+\frac{cos^2(pd)}{2} 
\end{pmatrix} 
\end{align*}

\begin{align}
\mathcal{M}_t & = e^{(-\frac{1}{2}(px - cp_{if})K(px - cp_{if})^T)}
\end{align}

\end{document}